\documentclass{article}
\usepackage{spconf,amsmath,graphicx} 

\usepackage{amssymb}
\usepackage[multiple]{footmisc}
\usepackage{tablefootnote}
\usepackage{xurl}
\usepackage{booktabs}
\usepackage{multirow}
\usepackage[normalem]{ulem}
\useunder{\uline}{\ul}{}
\usepackage[pagebackref,hidelinks]{hyperref}

\usepackage{algpseudocode}
\usepackage[ruled,vlined,linesnumbered]{algorithm2e}

\usepackage{pifont}
\newcommand{\xmark}{\text{\ding{55}}}
\newcommand{\cmark}{\text{\ding{51}}}

\newcommand{\dev}{{\fontfamily{qcr}\selectfont dev%
}}
\newcommand{\test}{{\fontfamily{qcr}\selectfont tst-COMMON%
}}

\def\email#1{\gdef\@email{#1}}
\gdef\@email{email@address}

 \hypersetup{
     colorlinks=true,
     linkcolor=blue,
     filecolor=blue,
     citecolor=blue,      
     urlcolor=cyan,
     }

\renewcommand*{\backref}[1]{}
\renewcommand*{\backrefalt}[4]{[{%
    \ifcase #1 Not cited.%
          \or #2.%
          \else #2.%
    \fi%
    }]}
    
\newcommand\blfootnote[1]{%
  \begingroup
  \renewcommand\thefootnote{}\footnote{#1}%
  \addtocounter{footnote}{-1}%
  \endgroup
}

\title{Efficient Speech Translation with Dynamic Latent Perceivers}

\twoauthors
    {Ioannis Tsiamas, 
     Gerard I. Gállego, 
     José A. R. Fonollosa
    }
	{Universitat Politècnica de Catalunya, Barcelona \\
	\small\texttt{\{ioannis.tsiamas,gerard.ion.gallego,jose.fonollosa\}@upc.edu}
	}
    {Marta R. Costa-jussà}
	{Meta AI, Paris \\
	{\small\texttt{costajussa@meta.com}}
	}

\begin{document}

\maketitle

\begin{abstract}
    Transformers have been the dominant architecture for Speech Translation in recent years, achieving significant improvements in translation quality. Since speech signals are longer than their textual counterparts, and due to the quadratic complexity of the Transformer, a down-sampling step is essential for its adoption in Speech Translation. Instead, in this research, we propose to ease the complexity by using a Perceiver encoder to map the speech inputs to a fixed-length latent representation. Furthermore, we introduce a novel way of training Perceivers, with Dynamic Latent Access (DLA), unlocking larger latent spaces without any additional computational overhead. Speech-to-Text Perceivers with DLA can match the performance of Transformer baselines across three language pairs in MuST-C. Finally, a DLA-trained model is easily adaptable to DLA at inference, and can be flexibly deployed with various computational budgets, without significant drops in translation quality.

\end{abstract}


\section{Introduction} \label{sec:intro}
\vspace{-0.1cm}

    Speech Translation (ST) has traditionally relied on a \emph{cascade} approach, using two separate systems, an Automatic Speech Recognition (ASR) for transcription and a Machine Translation (MT) for text translation. Recently, the \emph{end-to-end} approach, with a single model, has attracted more interest, having several advantages such as faster inference and no error propagation \cite{taking-stock,cascade-vs-e2e}. The Transformer \cite{attention-is-all-you-need} has been crucial for this change, becoming the standard model in end-to-end ST.
    
    \vspace{-0.43cm}
    
    \blfootnote{Work at UPC was supported by the Spanish State Research Agency (AEI) project PID2019-107579RB-I00 / AEI / 10.13039/501100011033.}
    
    \begin{figure}[ht]
        \centering
        \resizebox{0.93\columnwidth}{!}{
        \includegraphics{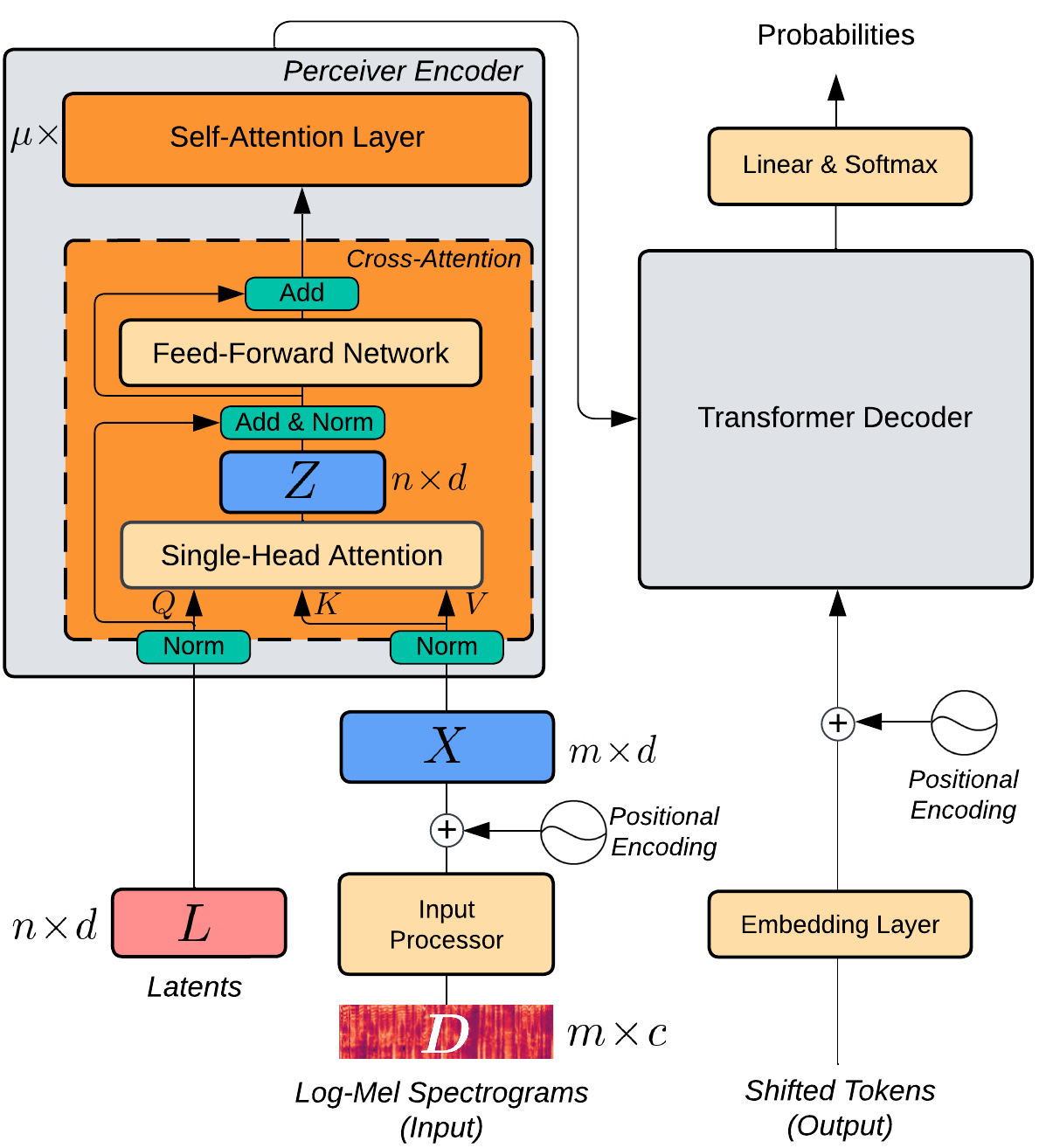}
        }
        \vspace{-0.1cm}
        \caption{Speech-to-Text Perceiver}
        \label{fig:model}
        \vspace{-0.3cm}
    \end{figure}
    
    One of the Transformer's key features is the ability to model token-to-token interactions with attention matrices, which imposes a quadratic complexity with respect to the sequence length. Since speech sequences are much longer than text sequences, directly processing speech with a Transformer becomes problematic. Thus, a modification is usually necessary, with down-sampling the speech signal at the input of the encoder \cite{s-transformer} or at the input of the attention modules \cite{speechformer}. In this research, we take an alternative approach and propose to map the input speech to a fixed-length latent representation using a Perceiver encoder \cite{perceiver}. This mapping swaps the quadratic complexity from the sequence length to the number of latents and makes the model only linearly dependent on the sequence length. We demonstrate that a Perceiver encoder coupled with a Transformer decoder can obtain competitive results across three language pairs in end-to-end ST. To further ease the computational burden of the proposed model, we introduce a novel way of training and doing inference with Perceivers, called Dynamic Latent Access (DLA). By enabling Perceivers to have access to a large latent space but only use a small part of it at each training step, we can increase the model's expressive power without incurring additional computational costs. We also show that a diversity-based DLA can be utilized during inference to achieve significant improvements in efficiency with minimal reduction in translation quality. Finally, we investigate the complementary nature of DLA at training and inference and show that combining the two can create a \emph{single} and \emph{flexible} model that can be used in various scenarios with varying computational budgets. Our code is publicly available.\footnote{\url{https://github.com/mt-upc/s2t-perceiver}}

\section{Relevant Research} \label{sec:relevant_research}
\vspace{-0.1cm}
    
    Many Transformer \cite{attention-is-all-you-need} variants have been proposed for speech tasks. They usually involve changing the encoder, by adding strided convolutional layers to down-sample the input \cite{s-transformer,speech-transformer}. Further variations include the introduction of convolution inside the attention layers \cite{speechformer,conformer}. In this work, we replace the encoder with a Perceiver \cite{perceiver}, enabling the model to work on a latent space with an arbitrary number of latents.
    
    The Perceivers \cite{perceiver,perceiverIO} is a family of attention-based encoders that do not depend on inductive biases and can thus be applied to different modalities with very few modifications. One of their key features is that they project the input to a fixed-length latent representation, alleviating the quadratic scaling problem of the Transformer \cite{attention-is-all-you-need}. The latents are learned parameters and their number is a hyperparameter, which remains fixed throughout training and inference. The Perceiver obtains competitive results on language understanding, image classification, and multimodal audio-video tasks. In this research, we take advantage of the scaling properties of the Perceiver to tackle Speech Translation, a sequence-to-sequence task that is characterized by long source sequences.
    
    The PerceiverAR \cite{perceiverAR} is an autoregressive decoder that uses the previous context as a latent initialization, and can thus allow for varying compute at inference time. On the contrary, our proposed method, DLA, selects latents \emph{dynamically} for each example and can be utilized at both training and inference time. Our method is also similar to techniques like LayerDrop \cite{layerdrop}, which helps in training deeper models without raising the computational costs. Instead of a deeper model, DLA allows training a Perceiver on large latent spaces that can be fully or partially used at inference time.

\vspace{-0.15cm}
\section{Proposed Methodology} \label{sec:methodology}
\vspace{-0.15cm}

    \setlength{\parindent}{0pt}
    
        \textbf{Architecture.} The Speech-to-Text Perceiver (Fig. \ref{fig:model}) employs a Perceiver encoder \cite{perceiverIO} coupled with a Transformer decoder \cite{attention-is-all-you-need}. The Perceiver encoder consists of an initial cross-attention layer, followed by several self-attention layers. The input to the Perceiver encoder is log-Mel spectrograms $D \in \mathbb{R}^{m \times c}$, where $m$ is the number of frames in the input and $c$ is the number of frequency bins. The input is first processed with a 2-layer non-strided convolutional network, followed by an addition of sinusoidal positional embeddings \cite{attention-is-all-you-need}, to obtain $X \in \mathbb{R}^{m \times d}$, where $d$ is the dimensionality of the model. A set of $n$ $d$-dimensional latent vectors $L \in \mathbb{R}^{n \times d}$ is also passed to the encoder. The latent vectors are parameters, that are randomly initialized and learned during training. The cross-attention layer uses a single-headed attention module \cite{attention-is-all-you-need} to map the latent vectors $L$ and the processed input $X$ to a latent representation $Z \in \mathbb{R}^{n \times d}$, which is then passed through a feed-forward network. Layer normalization \cite{layernorm} is applied to both the inputs $L$, $X$ of the attention, and to its output $Z$. Inputs to the attention and feed-forward modules are added residually to their outputs. The output of the cross-attention layer is then processed by $\mu$ self-attention layers \cite{attention-is-all-you-need} and passed to the Transformer decoder, which produces the output token probabilities.
    
        \begin{figure}[ht]
            \centering
            \resizebox{0.95\columnwidth}{!}{
            \includegraphics{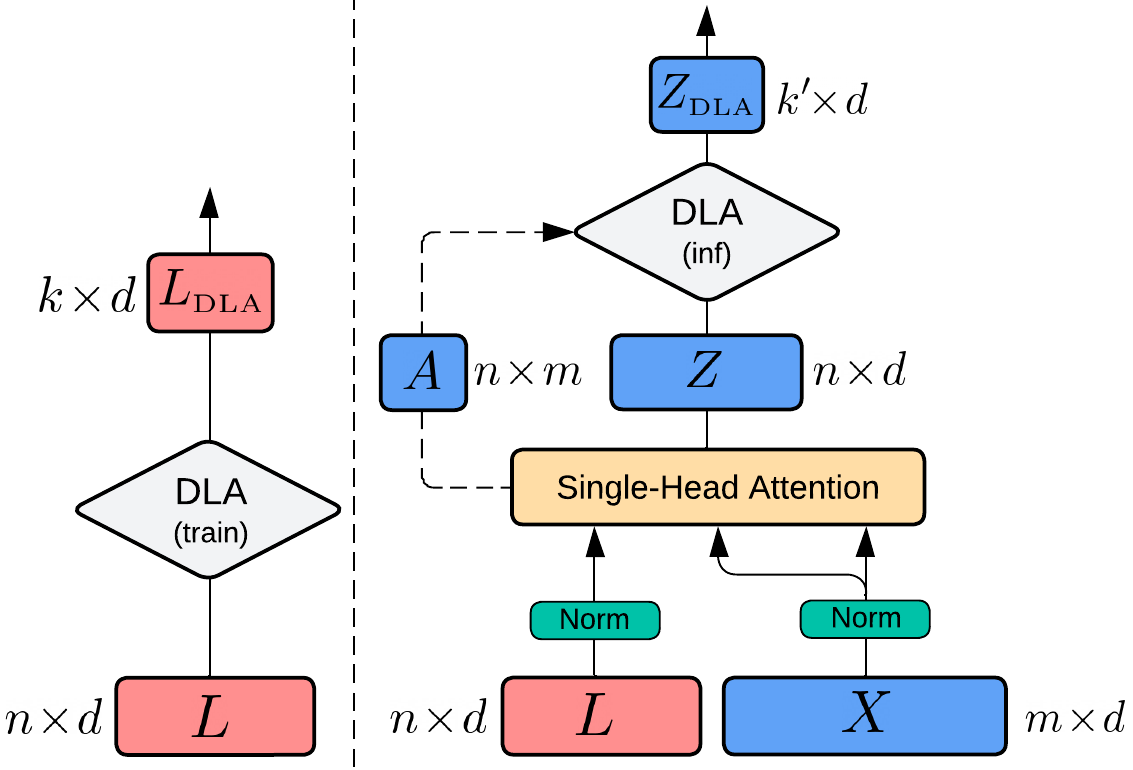}
            }
            \caption{Dynamic Latent Access. (a) Training (b) Inference}
            \label{fig:dla}
        \end{figure}
        
        \textbf{Dynamic Latent Access.} Since the input is mapped to a latent space of size $n$, the complexity with respect to the input length $m$ is only linear, i.e. $\mathcal{O}(nm)$, unlike the quadratic one of a Transformer encoder, $\mathcal{O}(m^2)$. This is a significant advantage, especially in the domain of ST, which is characterized by long input sequences that can even reach lengths of $ m=3,000$\footnote{Log-Mel filterbanks for a speech segment of 30 seconds.}. The size of the latent space $n$ is a hyperparameter, and in general higher values will provide more expressive power to the encoder. But due to the self-attention layers in the Perceiver encoder, there is now a quadratic complexity with respect to $n$. More specifically, $\mathcal{O}(nm + \mu n^2)$ for the whole encoder, where $\mu$ is the number of self-attention layers. Additionally, the choice of $n$ provides flexibility only once, before the training, and then the model is bound to it. To signify the benefits of the Perceiver encoder, we propose a novel way of utilizing the latent space, with Dynamic Latent Access (DLA). The proposed method can be used both at training ($\text{DLA}_\text{train}$, Fig. \hyperref[fig:dla]{2a}) and inference ($\text{DLA}_\text{inf}$, Fig. \hyperref[fig:dla]{2b}). At training time, DLA samples for each example randomly a set of $k$ latent vectors, $L_\text{DLA} \in \mathbb{R}^{k \times d}$, where $k \leq n$. Thus, $\text{DLA}_\text{train}$ can provide access to large latent space with size $n$, providing more capacity to the encoder, while being computationally bound only to $k$. DLA can also be used at inference to avoid the computationally expensive generation with $n$ latent vectors, in favor of $k'<n$. $\text{DLA}_\text{inf}$ is applied to the latent representation $Z$, and selects a set of $k'$ vectors $Z_\text{DLA} \in \mathbb{R}^{k' \times d}$, by maximizing the diversity of the corresponding attention weights $A \in \mathbb{R}^{n \times m}$ (Alg. \ref{alg:dla_inf}). We first calculate the absolute cosine similarity matrix $S \in \mathbb{R}^{n \times n}$ of the $\ell_2$-normalized $A$. Then, starting from the most diverse latent, we iteratively select latents up to $k'$, by minimizing the similarity score between the next latent and the most similar of the already selected ones. Since the attention weights are a function of both the latent space $L$ and the data $X$, they allow us to make a specialized selection for each example during inference. Note that Alg. \ref{alg:dla_inf} has a negligible computational burden, since $S$ is computed only once, and it is also batch-parallelizable.


    \vspace{-1ex}
    \begin{algorithm}[]
        \caption{\small{DLA - Inference}}
        
        \label{alg:dla_inf}
        \footnotesize
        
        \SetKwInOut{Input}{input}\SetKwInOut{Output}{output}
    
        \Input{$Z$ \Comment{Latent vectors, $n \times d$}}
        \Input{$A$ \Comment{Attention weights, $n \times m$}}
        \Input{$k'$ \Comment{$\#$ DLA-inf latents, integer}}
        \Output{$Z_\text{DLA}$ \Comment{DLA-inf latent vectors, $k' \times d$}}
        
        $I \gets \text{Empty List}$
        
        $\bar{A} \gets A / ||A||_2$ \Comment{$\ell_2$-normalized $A$, $n \times m$} 
        
        $S \gets |\bar{A}\bar{A}^T|$ \Comment{Absolute Similarity matrix, $n \times n$}
        
        $S \gets \text{Mask diagonal elements}$

        $i \gets \text{Select initial latent id}$

        $\text{append } i \text{ to } I$
    
        \While {$\text{len}(I) < k'$} {
    
            $\Sigma^* \gets S_{:,I}$ \Comment{$n \times \text{len}(I)$} 
            
            $scores \gets \text{max}(\Sigma^*)$ \Comment{$n \times 1$} 
            
            $scores \gets \text{Mask selected ids}$
            
            $i \gets \text{argmin}(scores)$ \Comment{Next most diverse latent id}
            
            $\text{append } i \text{ to } I$
        }

        $Z_\text{DLA} \gets Z_{I,:}$ \Comment{select the $I = [i_0, ..., i_{k'-1}]$ ids}
    
        \KwRet {$Z_\text{DLA}$} 
    \end{algorithm}
    \vspace{-1ex}

    \setlength{\parindent}{12pt}

\vspace{-0.15cm}
\section{Experimental Setup} \label{sec:exp_setup}
\vspace{-0.15cm}
    
    \setlength{\parindent}{0pt}
    
    \textbf{Data.} For our experiments we are using MuST-C \cite{mustc}, which is based on TED talks, and more specifically the pairs of English to German (En-De, 408 hours) from version 2.0, and the pairs of English to Spanish (En-Es, 504 hours), and English to Russian (En-Ru, 489 hours) from version 1.0.
    
    \textbf{Speech-to-Text Perceivers.} The Speech-to-Text Perceiver (S2T-Perceiver) models have 1 cross-attention layer and 12 self-attention layers in the encoder and 6 decoder layers, with dimensionality $d=256$. Apart from the Perceiver cross-attention, which is single-headed, 4 heads are used in the rest of the attention modules. The feed-forward layers have a hidden dimension of 2048 and GELU activations \cite{gelu}. Both the encoder and the decoder are using pre-LN \cite{preLN}. The latent array has the same dimensionality as the model (256) and is initialized with a truncated normal distribution with 0 mean and 0.05 standard deviation. A 2-layer non-strided convolutional network with 1024 inner channels, output dimensionality of 256, GLU activations \cite{glu} and kernel sizes of 5 process the 80-dimensional log-Mel spectrograms. Dropout of 0.15 is applied to all self-attention layers in the encoder and all layers in the decoder. Contrary to what is done usually in the Transformer \cite{attention-is-all-you-need,fairseq}, we found that is crucial for the training stability of the model to \emph{not} scale by $\sqrt{d}$ the processed input.
    
    \textbf{Baseline.} The Speech-to-Text Transformer (S2T-Transformer) has a similar architecture\footnote{We train the \textit{s2t\_transformer\_s} architecture from \textsc{fairseq} \cite{fairseq}.}. To achieve the same number of parameters with the S2T-Perceiver (32.5m), we use 13 encoder layers. We also use GELU activations. The 2-layer convolutional network has strides of 2 instead of 1, thus down-sampling the input by a rate of 4.
    
    \textbf{Training.} For training all the models we are using AdamW \cite{adamw} with a base learning rate of 0.002, a warm-up of 5,000 steps, and an inverse square root scheduler. We use gradient accumulation to scale the effective batch size to 512 examples. We use SpecAugment \cite{spec-augment} for data augmentation and label smoothing of 0.1. The target vocabularies are learned with SentencePiece \cite{sentencepiece} and have a size of 8,000. We stop training when performance does not improve for 15 consecutive epochs. The encoders are initialized from the same model configuration, pre-trained on the ASR part of the data. Models are implemented and trained with \textsc{fairseq} \cite{fairseq}.
    
    \textbf{Evaluation.} We average the 10 best checkpoints in the \dev \text{ } set and generate with a beam search of 5. Evaluation is done by measuring BLEU \cite{bleu} using sacreBLEU \cite{sacrebleu}. All experiments are repeated with 3 different seeds, and we report the average BLEU on \test.
    
    \setlength{\parindent}{12pt}

\vspace{-0.15cm}
\section{Results} \label{sec:results}
\vspace{-0.15cm}


    First, we experiment with S2T-Perceivers, with and without $\text{DLA}_\text{train}$, and compare them with S2T-Transformer baselines. Models without $\text{DLA}_\text{train}$ use $k \! = \! n$, while models with $\text{DLA}_\text{train}$ use larger $n$, and we set $k$ to $n/4$. In the upper part of Table \ref{tbl:main_results}, we observe that S2T-Perceivers achieve competitive results compared to the baseline, with an improvement in BLEU scores as the number of latents $n$ increases. In the lower part of Table \ref{tbl:main_results}, we observe further gains in all configurations when $\text{DLA}_\text{train}$ is used, without increasing the number of latents $k$ used during training. By applying $\text{DLA}_\text{train}$, S2T-Perceivers with $k \! = \! 512$ are capable of matching the baseline's performance on average, and surpass it for En-Ru. Furthermore, S2T-Perceivers with $k \! = \! 256$ are also competitive with the use of $\text{DLA}_\text{train}$, and reach a higher BLEU than S2T-Perceivers with $k \! = \! n \! = \! 512$, while being more efficient since they utilize half the number of latents during training. 

    \begin{table}[th]
        \centering
        \resizebox{0.9\columnwidth}{!}{
            \begin{tabular}{@{}lcccc@{}}
            \toprule
                                                               & \multicolumn{1}{l}{\textbf{En-De}} & \multicolumn{1}{l}{\textbf{En-Es}} & \multicolumn{1}{l}{\textbf{En-Ru}} & \textbf{Average} \\ \midrule
            \textbf{S2T-Transformer}                           & \textbf{24.4}                     & \textbf{28.0}                     & 15.4                               & \textbf{22.6}   \\ \midrule
            \textbf{S2T-Perceiver ($k \! = \! n$)}                             &                                    &                                    &                                     &                  \\
            $\,\,$ $k \! = \! n \! = \! 128$                                                & 22.4                              & 25.4                              & 14.1                               & 20.6            \\
            $\,\,$ $k \! = \! n \! = \! 256$                                                & 23.6                             & 26.8                              & 15.0                               & 21.8            \\
            $\,\,$ $k \! = \! n \! = \! 512$                                                & 24.0                              & 27.3                              & 15.3                               & 22.2            \\ \midrule
            \textbf{+} $\textbf{DLA}_\textbf{train}$ ($k < n$) &                                    &                                    &                                     &                  \\
            $\,\,$ $k \! = \! 128$ $\,\,$ $n \! = \! 512$                                         & 22.7                              & 26.4                              & 14.6                            & 21.2            \\
            $\,\,$ $k \! = \! 256$  $\,\,$ $n \! = \! 1024$                                     & 24.0                              & 27.7                              & 15.3                               & 22.3            \\
            $\,\,$ $k \! = \! 512$  $\,\,$  $n \! = \! 2048$                                       & \underline{24.2}                              & \underline{27.8}                              & \textbf{\underline{15.6}}                      & \textbf{\underline{22.6}}            \\ \bottomrule
            \end{tabular}
        }
        \vspace{-0.2cm}
        \caption{BLEU($\uparrow$) scores on \test. $n$ is the total number of latents. $k$ is the number of latents for $\text{DLA}_\text{train}$. \textbf{Bold} is best overall. {\ul Underlined} is best S2T-Perceiver.}
        \label{tbl:main_results}
    \end{table}
    
    Following, we apply $\text{DLA}_\text{inf}$ with $k'$ number of latents, and study its impact on the translation quality and efficiency (Table \ref{tbl:dla_inf}). To evaluate efficiency, we estimate the number of floating-point operations (FLOPS), with lower numbers indicating higher efficiency. For an S2T-Perceiver with varying $k'$ we estimate the total FLOPS required at inference time for \test\footnote{We do not consider batching and beam search.} and present them relatively to the ones required by the S2T-Transformer. We use the best configuration of the S2T-Perceiver, trained with $k \! = \! 512$ and $n \! = \! 2048$ (last row of Table \ref{tbl:main_results}). Our results indicate that although full inference with $k' \! = \! 2048$ (without $\text{DLA}_\text{inf}$) is very inefficient compared to the S2T-Transformer, we can scale down $k'$ substantially without significant losses in translation quality. Specifically, scaling $k'$ down to $n/8 \! = \! 256$, only results in a minor 0.1 point decrease in average BLEU, while it requires $0.85\times$ the FLOPS of the S2T-Transformer. We observe measurable drops in relative BLEU only when scaling $k'$ down to $n/16 \! = \! 128$, where BLEU decreases to $0.95\times$, but with the required FLOPS being further reduced to $0.59\times$.
    
    \begin{table}[th]
        \centering
        \resizebox{\columnwidth}{!}{
        \begin{tabular}{@{}lccccc@{}}
        \toprule
                                 & \multicolumn{4}{c}{\textbf{BLEU ($\uparrow$)}}                                                      & \multirow{2}{*}{\textbf{FLOPS ($\downarrow$)}} \\ 
                                 & \multicolumn{1}{l}{\textbf{En-De}} & \textbf{En-Es} & \textbf{En-Ru} & \textbf{Average}                 &                                                \\ \midrule
        \textbf{S2T-Transformer} & \textbf{24.4}                      & \textbf{28.0}  & 15.4           & \textbf{22.6 ($1.00\times$)} & $1.00\times$                                   \\ \midrule
        \textbf{S2T-Perceiver}   &                                    &                &                &                              &                                                \\
        $\,\, k' \! = \! 2048$           & 24.2                               & 27.8           & 15.6           & \textbf{22.6 ($1.00\times$)} & $5.50\times$                                   \\ \midrule
        $\,\, k' \! = \! 1024$           & 24.2                               & \textbf{28.0}  & 15.6           & \textbf{22.6 ($1.00\times$)} & $2.59\times$                                   \\
        $\,\, k' \! = \! 512$            & 24.2                               & 27.8           & \textbf{15.7}  & \textbf{22.6 ($1.00\times$)} & $1.39\times$                                   \\
        $\,\, k' \! = \! 256$            & 24.0                               & 27.7           & \textbf{15.7}  & 22.5 ($1.00\times$)          & $0.85\times$                                   \\
        $\,\, k' \! = \! 192$            & 23.8                               & 27.5           & 15.5           & 22.3 ($0.99\times$)          & $0.72\times$                                   \\
        $\,\, k' \! = \! 128$            & 23.2                               & 26.6           & 14.9           & 21.6 ($0.95\times$)          & $0.59\times$                                   \\
        $\,\, k' \! = \! 64$             & 18.5                               & 21.5           & 12.0           & 17.3 ($0.77\times$)          & \textbf{$0.47\times$}                          \\ \bottomrule
        \end{tabular}
        }
        \vspace{-0.2cm}
        \caption{$\text{DLA}_\text{inf}$ with $k'$ latents. BLEU scores and FLOPS on \test \text{} for the S2T-Perceiver ($k \! = \! 512$, $n \! = \! 2048$).}
        \label{tbl:dla_inf}
    \end{table}

\begin{figure}[th]
        \centering
        \includegraphics[width=0.9\columnwidth]{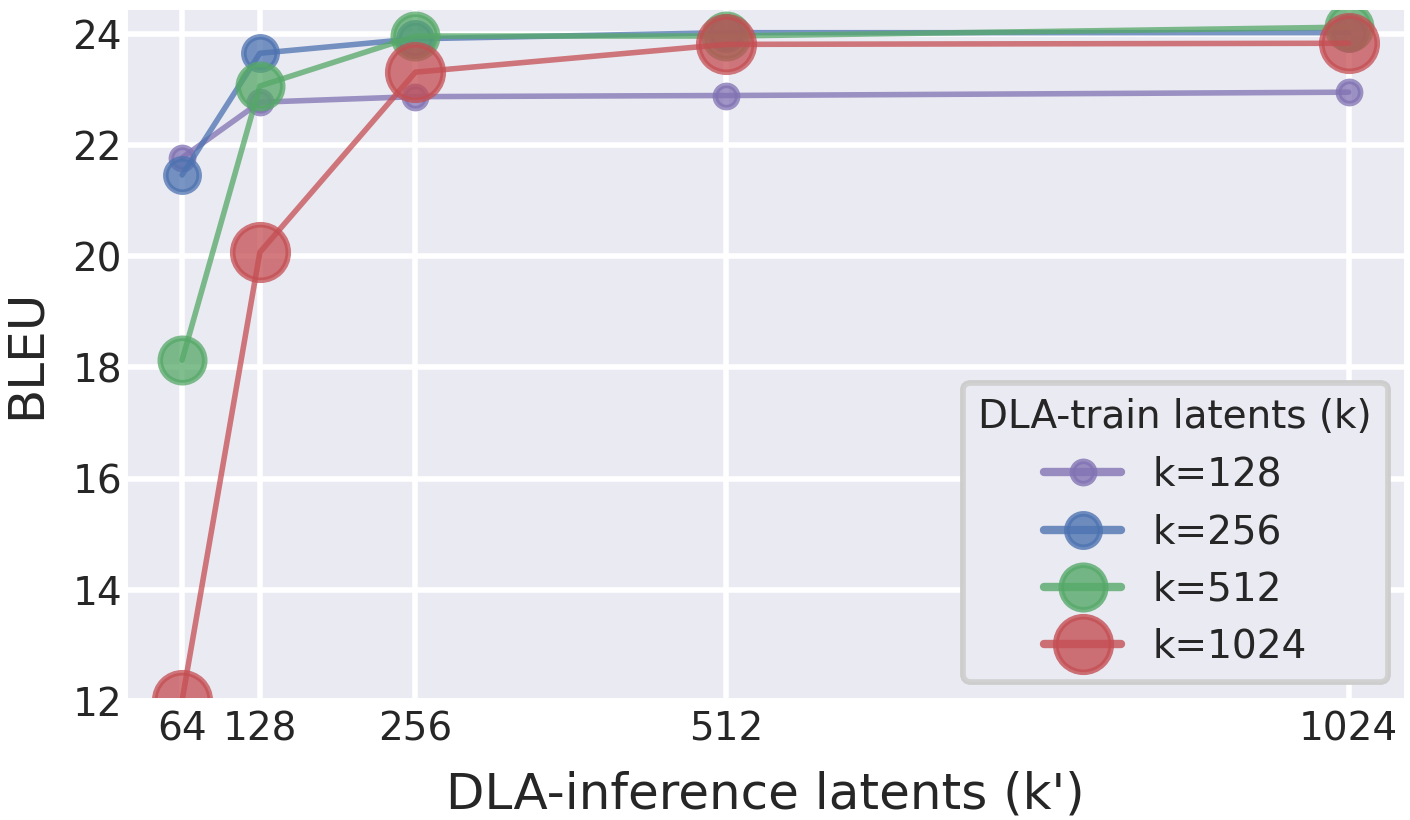}
        \vspace{-0.2cm}
        \caption{BLEU($\uparrow$) scores of four S2T-Perceivers ($n \! = \! 1024$, $k \! = \! 128, 256, 512, 1024$) on En-De \test \text{ } as a function of $\text{DLA}_\text{inf}$ latents ($k'$).}
        \label{fig:dla_inf}
    \end{figure}

    
    Next, we investigate the degree of compatibility between $\text{DLA}_\text{train}$ and $\text{DLA}_\text{inf}$. In Fig. \ref{fig:dla_inf} we compare four different S2T-Perceivers, which have access to the same number of latents $n \! = \! 1024$, but use different $\text{DLA}_\text{train}$ $k$ latents (128, 256, 512 and 1024). The configuration with $n \! = \! k \! = \! 1024$ essentially does not use $\text{DLA}_\text{train}$. For each model, we apply $\text{DLA}_\text{inf}$ with different values of $k'$ and report the BLEU scores on the En-De \test. We observe that the S2T-Perceiver without $\text{DLA}_\text{train}$ (red line) is not easily adaptable to a small number of inference latents $k'$, experiencing large drops in translation quality. On the other side, models with $\text{DLA}_\text{train}$ are much more compatible to $\text{DLA}_\text{inf}$, retaining most of their original BLEU scores for small values of $k'$. We also notice that training with few latents $k$, allows for better adaptability to $\text{DLA}_\text{inf}$, where the model with $k \! = \! 256$ only witnesses a drop in BLEU for an extremely small number of inference latents $k' \! = \! 64$. These findings indicate that $\text{DLA}_\text{train}$ does not only increases the performance with full inference, but also largely enables $\text{DLA}_\text{inf}$ for small values of $k'$. Finally, training with $k \! = \! 128$ also facilitates high adaptability but overall performance is sub-optimal, showing that no further gains are possible by setting $k$ to values below $n/4$.
    
    In the ablations of Table \ref{tbl:ablations} we find that not using a convolutional network to process the log-Mel spectrograms for the S2T-Perceiver, significantly lowers the translation quality. Contrary to \cite{perceiver}, we design a modality-specific architecture for a task suffering from data scarcity \cite{taking-stock,cascade-vs-e2e}, and thus we observe that introducing inductive biases through convolution is beneficial. Furthermore, we notice that down-sampling the sequence, results in a slightly worse performance, possibly due to information loss. Unlike the Transformer, the Perceiver can easily process the whole sequence, since it's not bound by its length. Finally, in Table \ref{tbl:ablations_dla} we compare the proposed $\text{DLA}_\text{inf}$, that maximizes latent diversity, with a version that selects latents randomly, and show the efficacy of the diversity-based selection, which is especially evident for smaller values of $k'$.
    
    \begin{table}[th]
        \centering
        \resizebox{0.9\columnwidth}{!}{
        \begin{tabular}{@{}cccccc@{}}
        \toprule
        \textbf{Input Proc.}  & \textbf{DS rate}   & \textbf{En-De} & \textbf{En-Es} & \textbf{En-Ru} & \textbf{Average} \\ \midrule
        $\cmark$ & $\boldsymbol{\times1}$ & \underline{24.2}          & \underline{27.8}          & \underline{15.6}          & \underline{22.6}            \\
        $\boldsymbol{\xmark}$ & $\boldsymbol{\times1}$ & 22.7          & 26.4          & 14.5          & 21.2 \\
        $\boldsymbol{\cmark}$ & $\boldsymbol{\times4}$ & 24.0         & 27.5          & 15.2          & 22.2           \\ \bottomrule
        \end{tabular}
        }
        \vspace{-0.2cm}
        \caption{Ablations on Input Processor and Down-Sampling. S2T-Perceiver ($k \! = \! 512$, $n \! = \! 2048$). BLEU on \test.}
        \label{tbl:ablations}
    \end{table}

    \vspace{-0.5cm}
    \begin{table}[th]
        \centering
        \resizebox{0.75\columnwidth}{!}{
        \begin{tabular}{@{}lcccccc@{}}
        \toprule
        $\boldsymbol{k'}$    & \textbf{64} & \textbf{128} & \textbf{192} & \textbf{256} & \textbf{512} & \textbf{1024} \\ \midrule
        \textbf{Diversity}    & {\ul 18.5}  & {\ul 23.2}   & {\ul 23.8}   & {\ul 24.0}   & {\ul 24.2}   & {\ul 24.2}    \\ 
        \textbf{Random} & 12.6        & 19.4         & 21.8         & 22.9         & 23.8         & {\ul 24.2}    \\ \bottomrule
        \end{tabular}
        }
        \vspace{-0.2cm}
        \caption{Diversity-based vs Random $\text{DLA}_\text{inf}$. S2T-Perceiver ($k \! = \! 512$, $n \! = \! 2048$). BLEU on En-De \test.}
        \label{tbl:ablations_dla}
    \end{table}

\vspace{-0.15cm}
\section{Conclusions} \label{sec:conclusions}
\vspace{-0.15cm}

    We presented a new paradigm for Speech Translation which relies on projecting the speech signal to an arbitrary-length latent space with a Perceiver. Furthermore, we introduced a method that allows the Perceiver to dynamically use part of a large latent space, boosting performance without additional costs. This also creates a single model that can flexibly operate on different computational budgets at inference time, with little loss in performance. Future research will take advantage of the proposed method's efficiency to model the much longer sequences required for context-aware Speech Translation.

\vfill\pagebreak


\bibliographystyle{IEEEbib}
\bibliography{references}

\end{document}